\documentclass[a4paper,conference]{IEEEtran}
\ifCLASSINFOpdf
   \usepackage[pdftex]{graphicx}
\else
\fi
\usepackage{amsmath}
\usepackage{amssymb}
\usepackage{array}

\ifCLASSOPTIONcompsoc
 \usepackage[caption=false,font=normalsize,labelfont=sf,textfont=sf]{subfig}
\else
 \usepackage[caption=false,font=footnotesize]{subfig}
\fi
\usepackage{url}

\usepackage{booktabs}       
\usepackage{xcolor}

\usepackage[some, placement=top]{background}

\backgroundsetup{
        placement=top,
        position={8cm,2cm}}

\SetBgScale{1}
\SetBgContents{\parbox{16cm}{\Large \centering This paper has been accepted for publication at the 25th International Conference on Pattern Recognition (ICPR), Milan, Italy, 2020. \textcopyright IEEE}}
\SetBgColor{gray}
\SetBgAngle{0}
\SetBgOpacity{0.9}

\begin{document}
%
\title{The Effect of Multi-step Methods on Overestimation in Deep Reinforcement Learning}

\author{\IEEEauthorblockN{Lingheng Meng}
\IEEEauthorblockA{Electrical and Computer Engineering\\
University of Waterloo\\
Waterloo, Canada\\
lingheng.meng@uwaterloo.ca}
\and
\IEEEauthorblockN{Rob Gorbet}
\IEEEauthorblockA{Knowledge Integration\\
University of Waterloo\\
Waterloo, Canada\\
rbgorbet@uwaterloo.ca}
\and
\IEEEauthorblockN{Dana Kuli\'c}
\IEEEauthorblockA{Monash University\\
Melbourne, Australia\\
dana.kulic@monash.edu}}


%


\maketitle
\BgThispage

\begin{abstract}
Multi-step (also called $n$-step) methods in reinforcement learning (RL) have been shown to be more efficient than the 1-step method due to faster propagation of the reward signal, both theoretically and empirically,  in tasks exploiting tabular representation of the value-function. Recently, research in Deep Reinforcement Learning (DRL) also shows that multi-step methods improve learning speed and final performance in applications where the value-function and policy are represented with deep neural networks. However, there is a lack of understanding about what is actually contributing to the boost of performance. In this work, we analyze the effect of multi-step methods on alleviating the overestimation problem in DRL, where multi-step experiences are sampled from a replay buffer. Specifically building on top of Deep Deterministic Policy Gradient (DDPG), we propose Multi-step DDPG (MDDPG), where different step sizes are manually set, and its variant called Mixed Multi-step DDPG (MMDDPG) where an average over different multi-step backups is used as update target of Q-value function. Empirically, we show that both MDDPG and MMDDPG are significantly less affected by the overestimation problem than DDPG with 1-step backup, which consequently results in better final performance and learning speed. We also discuss the advantages and disadvantages of different ways to do multi-step expansion in order to reduce approximation error, and expose the tradeoff between overestimation and underestimation that underlies offline multi-step methods. Finally, we compare the computational resource needs of Twin Delayed Deep Deterministic Policy Gradient (TD3), a state-of-art algorithm proposed to address overestimation in actor-critic methods, and our proposed methods, since they show comparable final performance and learning speed.
\end{abstract}


%
\IEEEpeerreviewmaketitle


\section{Introduction}

Deep Reinforcement Learning (DRL) \cite{mnih2015human} incorporates the powerful representation capacity of nonlinear Deep Neural Networks (DNNs) into classic Reinforcement Learning (RL), but also results in low data efficiency and instability, i.e., drastic fluctuations in accumulated reward increase rather than a smooth increase \cite{henderson2018deep,sunderhauf2018limits,rlblogpost}. Deep Q-Networks (DQNs) \cite{mnih2015human} represent the Q-value function with DNNs, but is only designed for discrete action spaces, while Deep Deterministic Policy Gradient (DDPG) \cite{lillicrap2015continuous} was proposed to  tackle continuous control tasks, where the actor and critic are both represented with DNNs. The move from low dimensional state and action spaces to high-dimensional state and action spaces by employing DNNs comes with the cost of low data efficiency and non-monotonic learning progress, where policy diverges from optimal due to inaccurate approximation of the value function. These disadvantages hinder DRL from broad use in applications where interactive data collection is time-consuming and smooth adaption of behavior is crucial to maintain engagement, e.g. interactive robots \cite{meng2019learning}.

Low data efficiency corresponds to slow learning speed, assuming the learning algorithm is capable of learning an optimal policy given sufficient data, and can be caused by two reasons: \textbf{1)} lack of data, and \textbf{2)} lack of training. If slow learning is caused by lack of data, the environment is under-explored.  In this case, an efficient exploration strategy, e.g. parameter space noise \cite{plappert2017parameter}, or a complementary source of experiences, e.g. World Models \cite{ha2018world}, can be helpful for generating additional training data. On the other hand, if slow learning is caused by lack of training, since enough experiences have been collected but not coded into policy, a more effective way to use the collected data is necessary such as prioritized replay buffer \cite{schaul2015prioritized,horgan2018distributed} and hindsight experience replay \cite{andrychowicz2017hindsight}.

Instability is partially related to the catastrophic forgetting problem of Deep Learning (DL) \cite{kirkpatrick2017overcoming} which is inherent in the continuously evolving nature of policy learning in Reinforcement Learning (RL). In addition, inaccurate estimation and continuous tuning of the Q-value function might lead the learned policy in directions far away from optimal or cause it to fluctuate around a local optimum. The overestimation problem \cite{thrun1993issues} in Q-learning is a typical example of inaccurate estimation in which the maximization of an inaccurate Q-value estimate induces a consistent overestimation. As the result of overestimation, the estimated Q-value of a given (state, action) pair might explode and drive the corresponding policy away from optimal. Therefore, DRL algorithms based on DNNs should strive to alleviate overestimation problem if it cannot be completely overcome.

Building on top of DDPG \cite{lillicrap2015continuous}, we experiment with Multi-step DDPG (MDDPG), where different step sizes are manually set, and with a variant called Mixed Multi-step DDPG (MMDDPG) where a mixture of different multi-step backups is used as target Q-value. In this paper, we refer backup target Q-values with more than 1 immediate reward as multi-step backups. We first experimentally show that MDDPG and MMDDPG outperform DDPG, in terms of final performance and learning speed, mainly because of their effect helps alleviate the overestimation problem. Then, we compare MMDDPG with other state-of-the-art approaches to show that the proposed method can achieve comparable performance to TD3 \cite{fujimoto2018addressing} which is a state-of-art DRL algorithm and dedicated to addressing overestimation problem. After that, we discuss the underestimation and overestimation underlying offline multi-step method. At the end, we conclude this work and provide prospects for future research.

\section{Related Work}
\label{sec:related_work}

Multi-step methods have been studied in traditional RL for both on- and off-policy learning considering both the forward view, i.e., updating each state by looking forward to future rewards and states, and the backward view, i.e., updating each state by combining the current Temporal Difference (TD) error with eligibility traces of past events \cite{sutton2018reinforcement}. Recently, a new multi-step action-value algorithm $Q(\sigma)$ was proposed to allow the degree of sampling performed by the algorithm at each step during its backup to be continuously varied, with Sarsa at one extreme, and Expected Sarsa at the other \cite{de2018multi}. The results show that an intermediate value of $\sigma$ performs better than either extreme. However, a systematic way to adjust $\sigma$ still needs to be studied, and the learning tasks in \cite{de2018multi} are relatively simple with small state and action spaces, avoiding the need for DL methods. Multi-step TD learning for non-linear function approximation was studied in \cite{van2016effective} where forward TD($\lambda$) was investigated on simple discrete control tasks and $\lambda$ is a hyper-parameter controlling the weight of different multi-step update targets. Although a neural network was used for function approximation in this work, only simple discrete control tasks were examined. Rainbow \cite{hessel2018rainbow}, an integrated learning agent combining many extensions including multi-step learning, found that multi-step not only helps speed up early learning but also improves final performance on Atari 2600 games. However, Rainbow is built on top of DQN and only discrete action space tasks were examined. In our work, we focus on continuous control tasks.

Multi-step methods have also been investigated in asynchronous methods \cite{mnih2016asynchronous}, which rely on parallel actors employing different exploration policies in parallel instances of environments, to decorrelate consecutive updating experiences and to stabilize policy learning without using a replay buffer. However, such a parallel paradigm can only work practically in simulated environments, and is unfeasible for real applications where multiple instances of physical systems (e.g., robots) are too expensive.

The overestimation problem \cite{thrun1993issues} is cited as the reason non-linear function approximation fails in RL. Based on Double Q-learning \cite{hasselt2010double}, Double DQN \cite{van2016deep} was shown to be effective in alleviating this problem for discrete action spaces.  Although TD3 \cite{fujimoto2018addressing} proposed to take the minimum of two  bootstrapped Q-values of a state-action pair which are separately estimated by two critics to overcome the overestimation problem, the extra neural network for the second critic also introduces additional computation cost, especially when the state and action spaces are large. 

Model-based Value expansion (MVE) \cite{feinberg2018model} is a multi-step method that expands multi-step on a learned environment model, whose performance tends to degrade in complex environments. As an improvement of MVE, STochastic Ensemble Value Expansion (STEVE) \cite{buckman2018sample} expands various multi-steps on an ensemble of learned environment models, including transition dynamics and reward function, then adds these to an ensemble of Q-value functions. The final target Q-value is a weighted mean of these generated target Q-values. Both MVE and STEVE suffer from modeling error, and more importantly they both introduce vast extra computation, especially STEVE.

Averaged-DQN \cite{anschel2017averaged} is proposed to reduce variance and stabilize learning by exploiting an average over a set of target Q-values calculated from a set of past Q-value functions. It is shown that Averaged-DQN also helps in alleviating the overestimation problem. Different from that, our method MMDDPG takes the average over a set of target Q-values calculated with different step sizes, which is also shown to be more stable than MDDPG.

\section{Background}

\subsection{Reinforcement Learning}
An RL agent learns an optimal policy through interactions with the external environment. Formally, RL can be formulated as a Markov Decision Process (MDP), where optimal actions are chosen by the agent to maximize the expected reward over discrete time-steps in an environment. An MDP is defined by a tuple $(S, A, p, r)$ with state space $S$, action space $A$, transition probability $p$, and reward function $r$. At each discrete time step $t$, an agent selects an action $a_t \in A$ at state $s_t \in S$ following its policy $\pi(a_t|s_t)$.  As a result of the action, the agent transitions to the next state $s_{t+1} \in S$ according to transition probability $p\left ( s_{t+1} | s_t, a_t \right ) \in [0,1]$ and receives a reward $r\left ( s_t, a_t, s_{t+1} \right ) \in [r_{min},r_{max}]$. The goal of the agent is to learn an optimal policy $\pi^{*}$ which maximizes the expectation of a discount accumulated reward $\mathbb{E}_{\pi^{*}, s_0 \sim \rho_0 }\left [ \sum_{i=0}^{\infty}\gamma^{i}r_{i} | s_0 \right ]$ where $\gamma$ is the discount factor and $\rho_0$ is the distribution of the initial state $s_0$.

Almost all RL algorithms estimate the state-value function V or the action value function Q. A V-value function of a state $s$ under a policy $\pi$ is the expected return when starting in $s$ and following $\pi$ thereafter, which can be formally defined by $ V^{\pi }\left ( s_t \right ) = \mathbb{E}_{\pi}\left [ \sum_{i=t}^{\infty } \gamma^{i-t}r_{t} | s_t \right ]$. A Q-value function represents the expected accumulated future reward when taking action $a_t$ in state $s_t$ and following a policy $\pi$ afterwards, and can be formally expressed as $Q^{\pi}\left ( s_t, a_t \right ) = \mathbb{E}_{\pi}\left [ \sum_{i=t}^{\infty} \gamma^{i-t} r_{t} | s_t, a_t \right ]$.

\subsection{Multi-step Bootstrapping}
\label{subsec:multi_step_bootstrapping}

The motivation of Multi-step methods \cite{sutton2018reinforcement} is to facilitate fast propagation of knowledge about the outcomes of selected actions, by taking into account rewards by looking $n$ steps forward. Formally, $n$-step discounted accumulated reward can be expressed as 
\begin{equation}
    R_{t}^{\left ( n \right )} = \sum_{i=0}^{n-1} \gamma ^{i} r_{t+i} ,
    \label{eq:n_step_return}
\end{equation} 
where when $n=1$ only one reward is considered, while for $n>1$ all rewards received between $t$ and $t+n-1$ are considered. Based on this $n$-step discounted accumulated reward, the corresponding Bellman optimality equation for $Q^{*}$ and $V^{*}$ \cite{hernandez2019understanding} can be defined as 
\begin{equation}
    Q^{*}\left ( s_t, a_t \right ) = \mathbb{E}_{\pi ^{*}}\left [ R_{t}^{*\left ( n \right )} + \gamma ^{n} \max_{a} Q^{*}\left ( s_{t+n}, a \right ) \right ],
\end{equation}
 and 
\begin{equation}
    V^{*}\left ( s_t \right ) = \mathbb{E}_{\pi ^{*}}\left [ R_{t}^{*\left ( n \right )} + \gamma ^{n} V^{*}\left ( s_{t+n}\right ) \right ]\text{,}
\end{equation}
where $R_{t}^{*\left ( n \right )}$ is the $n$-step discounted accumulated reward following optimal policy $\pi^{*}$. 

\subsection{Deep Deterministic Policy Gradient (DDPG)}
DDPG uses an actor-critic framework, implemented by deep neural networks, where the policy update is based on the Deterministic Policy Gradient \cite{silver2014deterministic}. In DDPG, the critic, $Q_{\theta^{Q}}\left ( s_t,a_t\right )$, is a Q-value function approximated by a deep neural network with weights $\theta^{Q}$ and is updated following Eq. \ref{eq:ddpg_critic}:
\begin{equation}
    L_{\theta^{Q}} = \mathbb{E}_{\left ( s_t,a_t,r_t,s_{t+1} \right )\sim U\left ( D \right )}\left [ \left ( \hat{Q}_{t} - Q_{\theta^{Q}}\left ( s_t,a_t\right ) \right )^2 \right ],
    \label{eq:ddpg_critic}
\end{equation}
where $\hat{Q}_{t}=r_t + \gamma Q_{\theta ^{Q-}}\left ( s_{t+1},\mu_{\theta ^{\mu -}} \left ( s_{t+n} \right ) \right )$, $Q_{\theta^{Q-}}$ and $\mu_{\theta ^{\mu -}} $ are the target critic and actor, parameterized by $\theta^{Q-}$ and $\theta ^{\mu -}$ respectively, which are soft-updated following Eq. \ref{eq:ddpg_target_networks_update}:
\begin{equation}
    \theta ^{Q-} \leftarrow  \left ( 1-\tau \right )\theta ^{Q-} + \tau\theta ^{Q} \ \text{,} \ \ \ \ \theta ^{\mu-} \leftarrow  \left ( 1-\tau \right )\theta ^{\mu-} + \tau\theta ^{\mu}
    \label{eq:ddpg_target_networks_update}
\end{equation}
with $0 < \tau\ll 1$ to stabilize the learning target, and experiences $\left ( s_t,a_t,r_t,s_{t+1} \right )$ are randomly sampled from replay buffer $D$ as indicated as a uniform distribution $U\left ( \cdot  \right )$. In DDPG, the actor, $\mu_{\theta ^{\mu }}\left ( s_t \right)$, is a deterministic policy represented by another deep neural network with weights $\theta ^{\mu }$ and is updated following Eq. \ref{eq:ddpg_actor}: 
\begin{equation}
    \centering
    \begin{aligned}
    \nabla_{\theta^{\mu}} J & \approx \mathbb{E}_{s_t\sim U\left ( D \right )}\left [ \nabla_{\theta^{\mu}} Q_{\theta ^{Q}}\left ( s_t, a  \right ) \right ] \\
                                      & = \mathbb{E}_{s_t\sim U\left ( D \right )}\left [ \nabla_{a} Q_{\theta ^{Q}}\left ( s_t, a  \right ) \nabla_{\theta^{\mu}}\mu_{\theta^{\mu }} \left ( s_t \right ) \right ],
    \end{aligned}
    \label{eq:ddpg_actor}
\end{equation}
where $a= \mu_{\theta ^{\mu }}\left ( s_t \right)$. All experiences are collected by adding exploratory noise to the deterministic policy, $a'= \mu_{\theta ^{\mu }}\left ( s_t \right)+\epsilon_a$.

\section{Proposed Methods}

In this section, we will progressively introduce MDDPG and MMDDPG. 

\subsection{Multi-step DDPG (MDDPG)}

MDDPG is a variant of DDPG  where multi-step experiences sampled from the replay buffer are used to calculate the direct accumulated reward, which is then added to the bootstrapped Q-value after these experiences. 

Based on the $n$-step discounted accumulated reward in Eq. \ref{eq:n_step_return}, we can easily realize $n$-step bootstrapped return using consecutively stored experiences in the replay buffer. Assuming that past experiences of an agent are consecutively stored in the replay buffer $D$ and the experience at time step $t$ is sampled into a training mini-batch, the $n$ consecutive experiences from $t$ to $t+n-1$ are treated as a single multi-time step sample. Then, for each sample in the  $n$-step mini-batch $\left \{ (s_t,a_t,r_t, \cdots, s_{t+n}, d_{t+n})^{\left ( i \right )} \right \}_{i=1}^{N}$ with size $N$, the $n$-step bootstrapped estimated action value function can be defined as Eq. \ref{eq:n_step_bootstrapped_backup_target}:
\begin{equation}
    \begin{aligned}
        \hat{Q}_{t}^{\left ( n \right )} = 
                    \left\{\begin{matrix}
                        &\sum_{i=0}^{n-1}{\gamma^{i}r_{t+i} } + \gamma^{n}\max_{a} Q_{\theta ^{Q-}}\left ( s_{t+n},a \right ),  \quad \quad \quad \quad \quad  \\ 
                        & if \ \forall \ k\in\left [ 1,\cdots ,n \right ] \ and \ d_{t+k}\neq 1;\\ 
                        &\sum_{i=0}^{k-1}{\gamma^{i}r_{t+i} }, \qquad \qquad \qquad \qquad \qquad \qquad \qquad \qquad  \ \\ 
                        & if \ \exists \ k\in\left [ 1,\cdots ,n \right ] \ and \ d_{t+k}=1.
                    \end{matrix}\right.
    \end{aligned}
    \label{eq:n_step_bootstrapped_backup_target}
\end{equation}
with $d_{t+k}=1$ if the episode is done, otherwise $d_{t+k}=0$.

The value function $Q$ of MDDPG is updated by minimizing the objective given in Eq. \ref{eq:n_step_Q_update_objective}:
\begin{equation}
    \begin{split}
        L_{\theta ^{Q}} =\mathbb{}_{(s_t,a_t,r_t, \cdots, s_{t+n}, d_{t+n})\sim U\left(D\right)} \left [ \left( \hat{Q}_{t}^{\left ( n \right )} - Q_{\theta ^{Q}} \left ( s_t, a_t  \right ) \right )^{2} \right ],
    \end{split}
    \label{eq:n_step_Q_update_objective}
\end{equation}
where $\hat{Q}_{t}^{\left ( n \right )}$ is Eq. \ref{eq:n_step_bootstrapped_backup_target}, while the policy update remains the same as DDPG. For convenience, in this paper we will denote MDDPG with  step size $n$ as MDDPG($n$). Specifically, when $n=1$, MDDPG($1$) is equivalent to DDPG.

\subsection{Mixed Multi-step DDPG (MMDDPG)}
MMDDPG is a variant of MDDPG based on the observation that for different tasks the best choice of step size $n$ may differ. MMDDPG mixes target Q-values calculated with different step sizes. This also helps to reduce the bias of the target Q-value by mixing a small set of target Q-values. The mixture can be an average over target Q-values with different step sizes from $1$ to $n$ as $\hat{Q}_{t}^{\left ( n_{avg} \right )}$ in Eq. \ref{eq:mixed_multi_step_ddpg}, or the minimum of such a set of target Q-values as $\hat{Q}_{t}^{\left ( n_{min} \right )}$ in Eq. \ref{eq:mixed_multi_step_ddpg}. Or considering $n=1$ is the most prone to overestimation, MMDDPG could take the average over target Q-values with step size from $2$ to $n$, as $\hat{Q}_{t}^{\left ( n_{avg-1} \right )}$ in Eq. \ref{eq:mixed_multi_step_ddpg}: 

\begin{equation}
    \begin{aligned}[b] 
        \hat{Q}_{t}^{\left ( n_{avg} \right )} = \frac{1}{n}\sum_{i=1}^{n}\hat{Q}_{t}^{\left ( i \right )} \ \ \ \ &\text{or} \  \ \ \ \hat{Q}_{t}^{\left ( n_{min} \right )} = \min_{i\sim \left [ 1,n \right ]}\hat{Q}_{t}^{\left ( i \right )} \\ 
        \ \text{or} \ \ \ \hat{Q}_{t}^{\left ( n_{avg-1} \right )} &= \frac{1}{n-1}\sum_{i=2}^{n}\hat{Q}_{t}^{\left ( i \right )}.
    \end{aligned}
    \label{eq:mixed_multi_step_ddpg}
\end{equation}

Similar to MDDPG($n$), we will denote MMDDPG with different mixture methods introduced in Eq. \ref{eq:mixed_multi_step_ddpg} as MMDDPG($n$-avg), MMDDPG($n$-min), and MMDDPG($n$-avg-1), respectively.

\section{Experiments}

\begin{figure*}[!t]
    \centering
    \includegraphics[width=.95\linewidth]{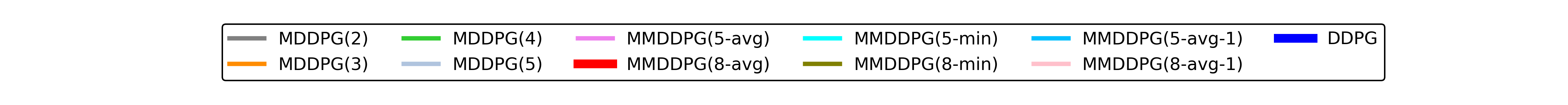}
    \subfloat[AntPyBulletEnv-v0]{\includegraphics[width=.5\linewidth]{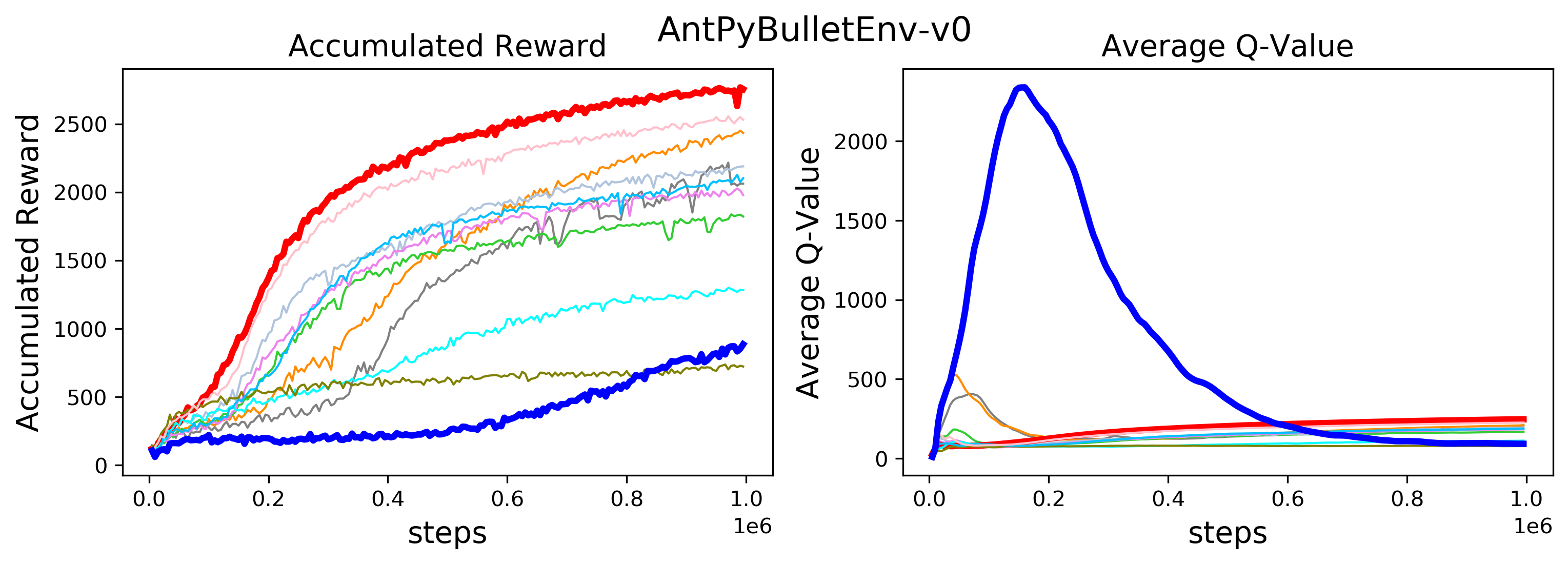}%
    \label{fig_first_case}}
    \hfil
    \subfloat[HalfCheetahPyBulletEnv-v0]{\includegraphics[width=.5\linewidth]{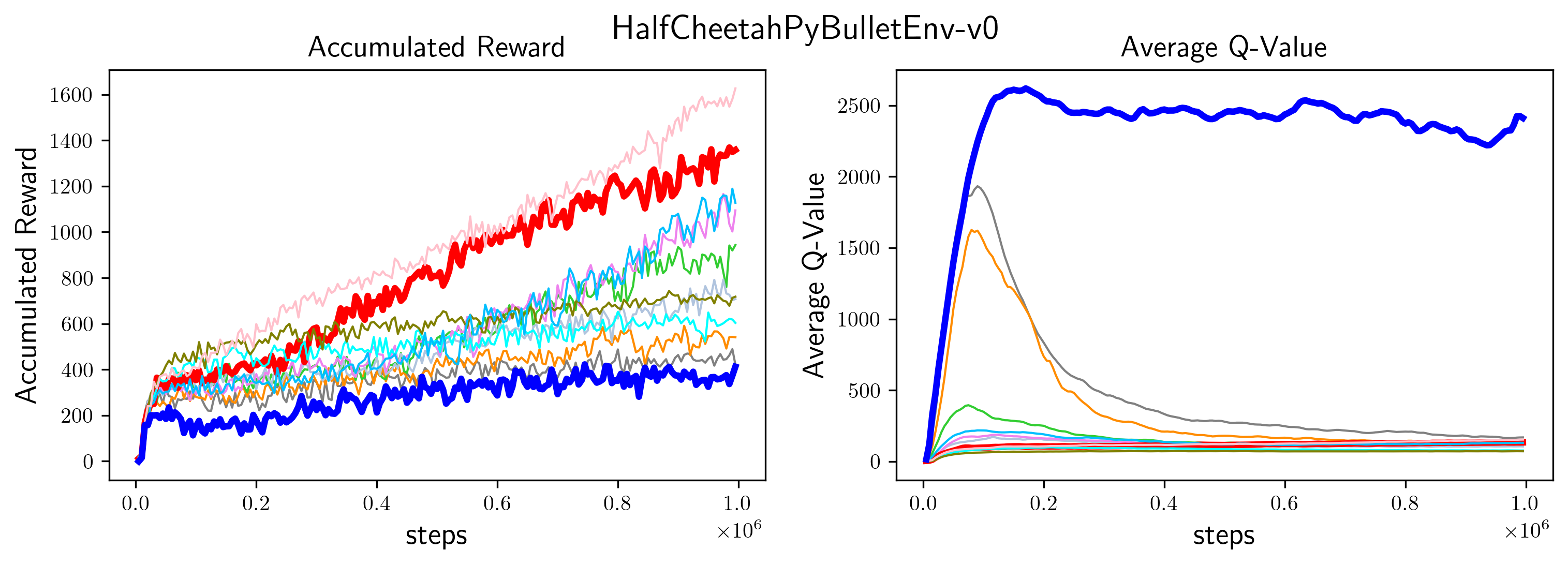}%
    \label{fig_second_case}}
    
    \subfloat[HopperPyBulletEnv-v0]{\includegraphics[width=.5\linewidth]{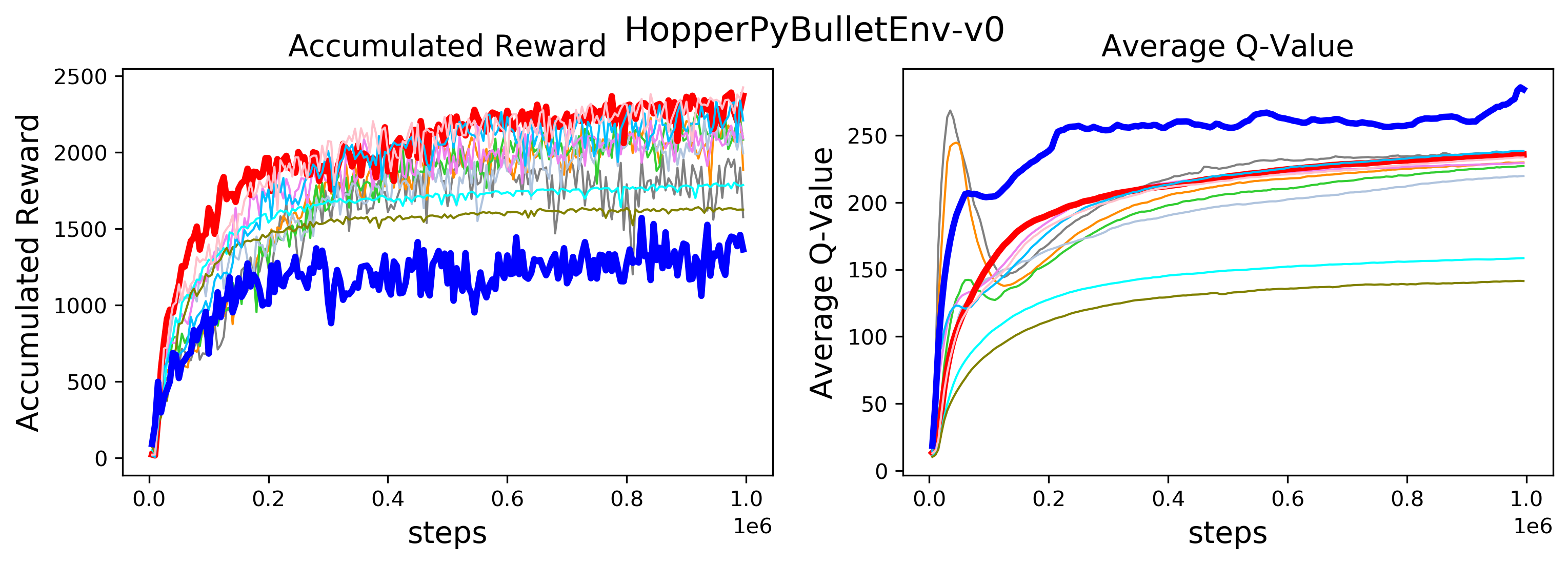}%
    \label{fig_second_case}}
    \subfloat[Walker2DPyBulletEnv-v0]{\includegraphics[width=.5\linewidth]{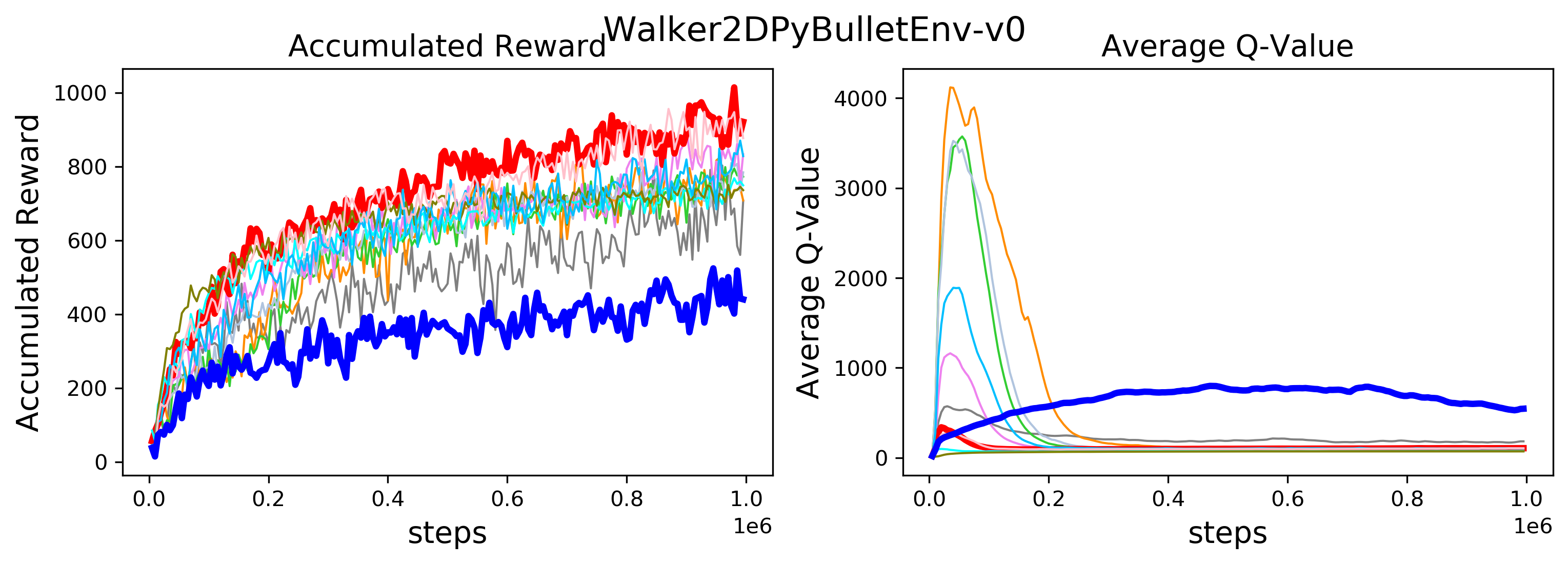}%
    \label{fig_second_case}}
    
    \caption{Comparison Among MDDPG, MMDDGP and DDPG, where for each task accumulated reward and average Q-value are shown side-by-side correspondingly to demonstrate the relationship between the overestimation of Q-value and performance.}
    \label{fig:Comparison_Among_MDDPG_MMDDGP_and_DDPG}
\end{figure*}

Tasks used for evaluation are all continuous control tasks from PyBulletGym\footnote{\url{https://github.com/benelot/pybullet-gym}} which provides OpenAI Gym \cite{1606.01540} compatible environments based on open-source Bullet Physics engine rather than non-free MuJoCo engine. We compare MDDPG and MMDDPG with vanilla DDPG, TD3, SAC \cite{haarnoja2018soft}, MVE and STEVE.\footnote{DDPG, TD3 and SAC use implementations in \url{https://github.com/openai/spinningup}, and MVE and STEVE use the implementations in \url{https://github.com/tensorflow/models/tree/master/research/steve}} All algorithms use policy and value functions with 2 hidden layers and each layer has 300 hidden units. Other hyper-parameters are set to the default values. All experiments are run five times for five different random seeds.

\subsection{Experimental Evidence of Multi-step Methods' Effect on Alleviating Overestimation}
\label{subsec:Experimental_Evidence_of_Multi_step_Methods_Effect_on_Alleviating_Overestimation}

Fig. \ref{fig:Comparison_Among_MDDPG_MMDDGP_and_DDPG} compares DDPG with its variants MDDPG and MMDDPG with different step size $n$. This figure illustrates that all MDDPG($n$) with $n>1$ outperform DDPG, and especially for MMDDPG($8$-avg) the improvement, in terms of final performance and learning speed, is significant as highlighted with the red line. To illustrate the underlying relationship between the performance and learned Q-value, the average Q-value is shown in parallel, from which we can see that the bad performance of DDPG always corresponds to an extremely large Q-value. Note that even though multi-step methods help to relieve the overestimation problem, they cannot completely overcome this problem, as shown by the drastic increase and followed by the sharp decrease in Q-value within the first few epochs in Fig. \ref{fig:Comparison_Among_MDDPG_MMDDGP_and_DDPG}, whereas DDPG takes more time to decrease its Q-value, and in some cases never does. The initial overestimation is caused by approximation error on most (state, action) pairs, because at the beginning stage of the learning only a small set of (state, action) pairs are encountered, causing a high error in (state, action) pairs without training data. However, as more and more experiences are collected in the replay buffer, the approximation error is reduced. From the average Q-values in Fig. \ref{fig:Comparison_Among_MDDPG_MMDDGP_and_DDPG}, we can see that none of the examined approaches avoid this initial explosion in the average Q-values.

   \begin{figure}[thpb]
      \centering
      \includegraphics[width=.99\linewidth]{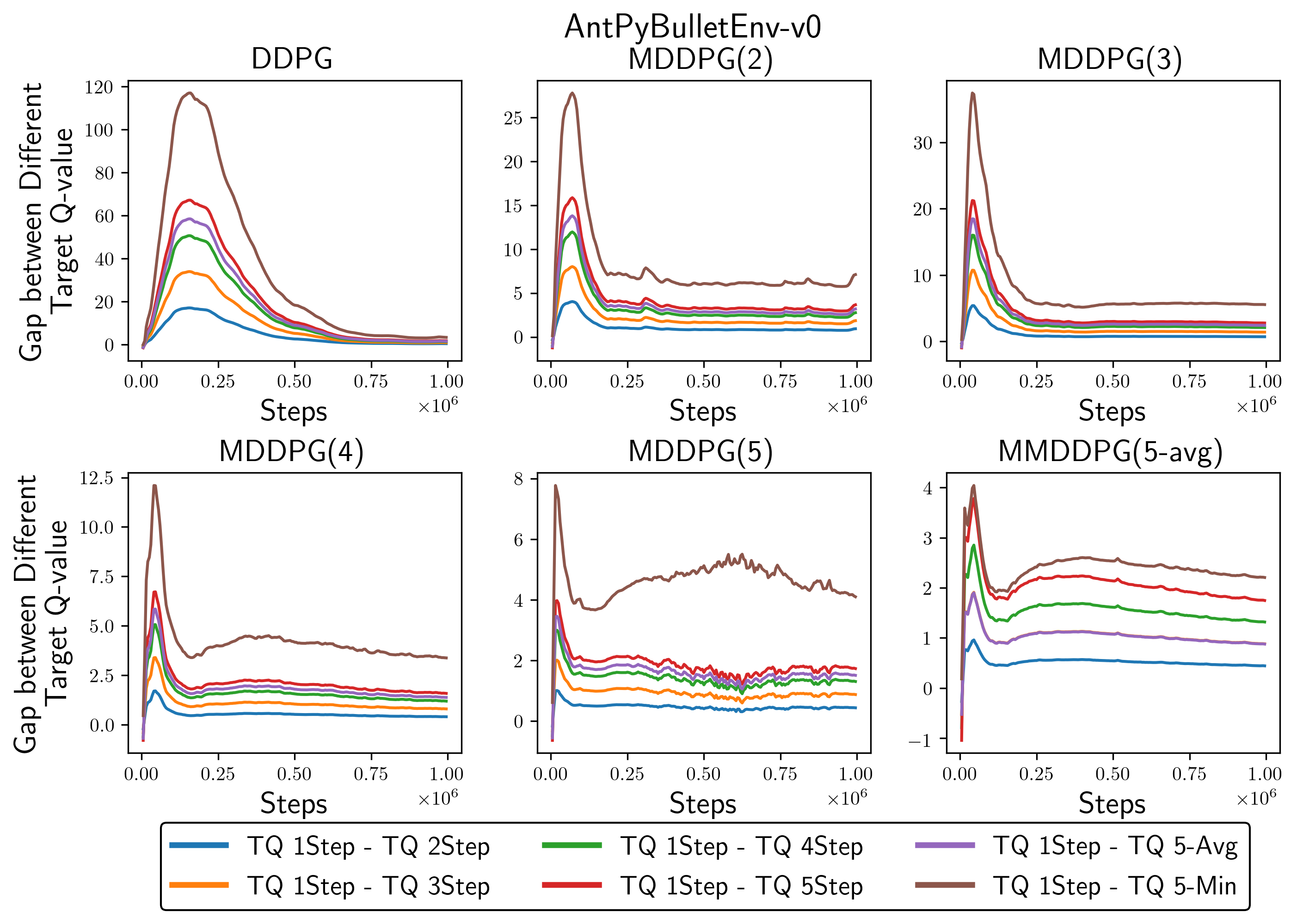}
      \caption{The Difference in Estimated Target Q-values Between 1-step and Multi-step Methods, where the larger the value, the bigger the difference. (MDDPG($n$) is a multi-step DDPG with step size $n$, and MMDDPG($n$-avg) is a mixture of $1$- to $n$-step DDPG.)}
      \label{fig:backup_gaps}
   \end{figure}

To investigate why multi-step methods help to alleviate the overestimation problem, we record backups of sampled experiences $\hat{Q}_{t}^{\left ( 1 \right )}$, $\hat{Q}_{t}^{\left ( 2 \right )}$, $\hat{Q}_{t}^{\left ( 3 \right )}$, $\hat{Q}_{t}^{\left ( 4 \right )}$, $\hat{Q}_{t}^{\left ( 5 \right )}$, $\hat{Q}_{t}^{\left ( n_{avg} \right )}$, and $\hat{Q}_{t}^{\left ( n_{min} \right )}$ for DDPG, MDDPG($n$) and MMDDPG($n$-avg) to depict the gap between 1-step and multi-step backups. For example, $\hat{Q}_{t}^{\left ( 1 \right )} - \hat{Q}_{t}^{\left ( 2 \right )}$, indicated as \textit{``TQ 1Step- TQ 2Step"} in Fig. \ref{fig:backup_gaps} shows the difference between 1-step and 2-step backups. Four key characteristics can be observed in this figure: \textbf{(1)} within a specific algorithm all gaps are positive which means multi-step methods provide smaller estimated target Q-values than that of the $1$-step method; \textbf{(2)} the larger the step, the smaller the corresponding estimated target Q-value, e.g. the blue line underneath the yellow line in each sub-figure; \textbf{(3)} the difference becomes smaller with increased interactions; and \textbf{(4)} among the different algorithms, the magnitude of the estimated Q-value decreases as the step size $n$ increases. These findings provide insight into multi-step methods' effect on alleviating the overestimation problem.

\subsection{Performance Comparison}
\label{subsec:Performance_Comparison}

\begin{figure*}[!t]
    \centering
    \includegraphics[width=.95\linewidth]{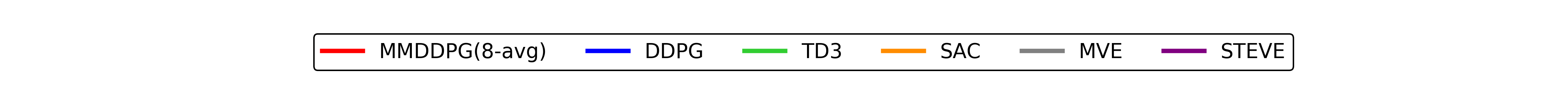}
    \subfloat[AntPyBulletEnv-v0]{\includegraphics[width=.25\linewidth]{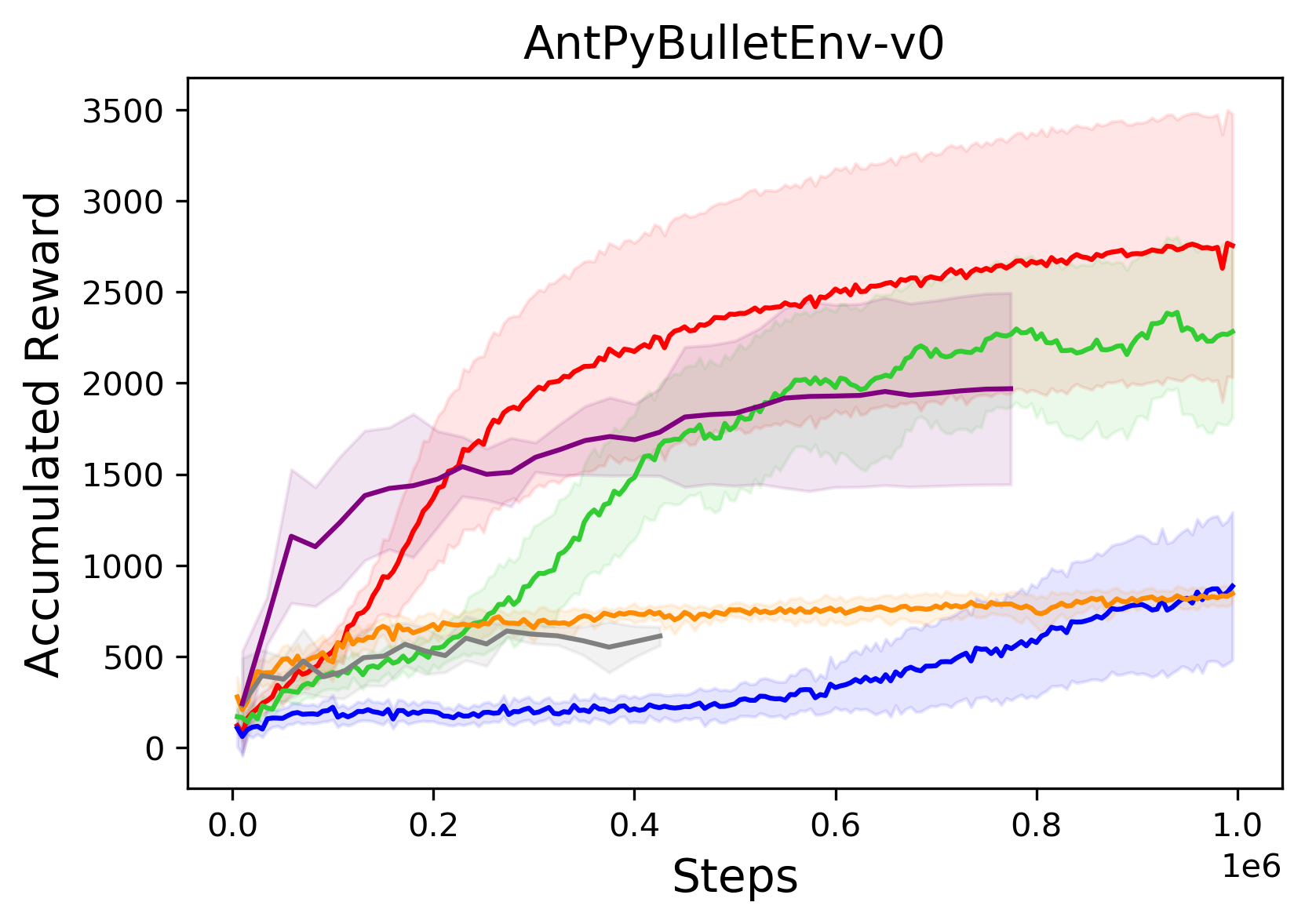}%
    \label{fig_first_case}}
    \hfil
    \subfloat[HalfCheetahPyBulletEnv-v0]{\includegraphics[width=.25\linewidth]{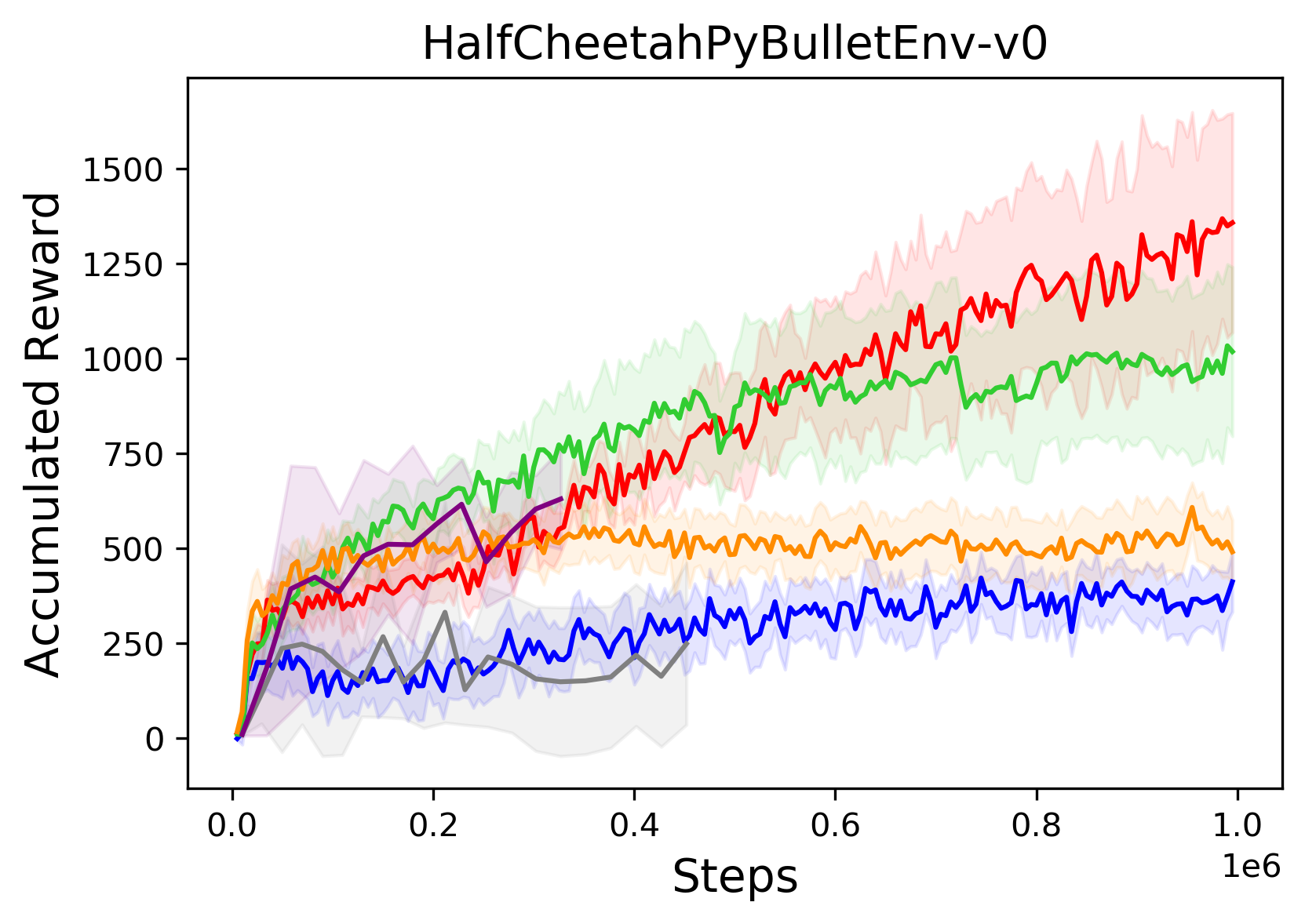}}
    \subfloat[HopperPyBulletEnv-v0]{\includegraphics[width=.25\linewidth]{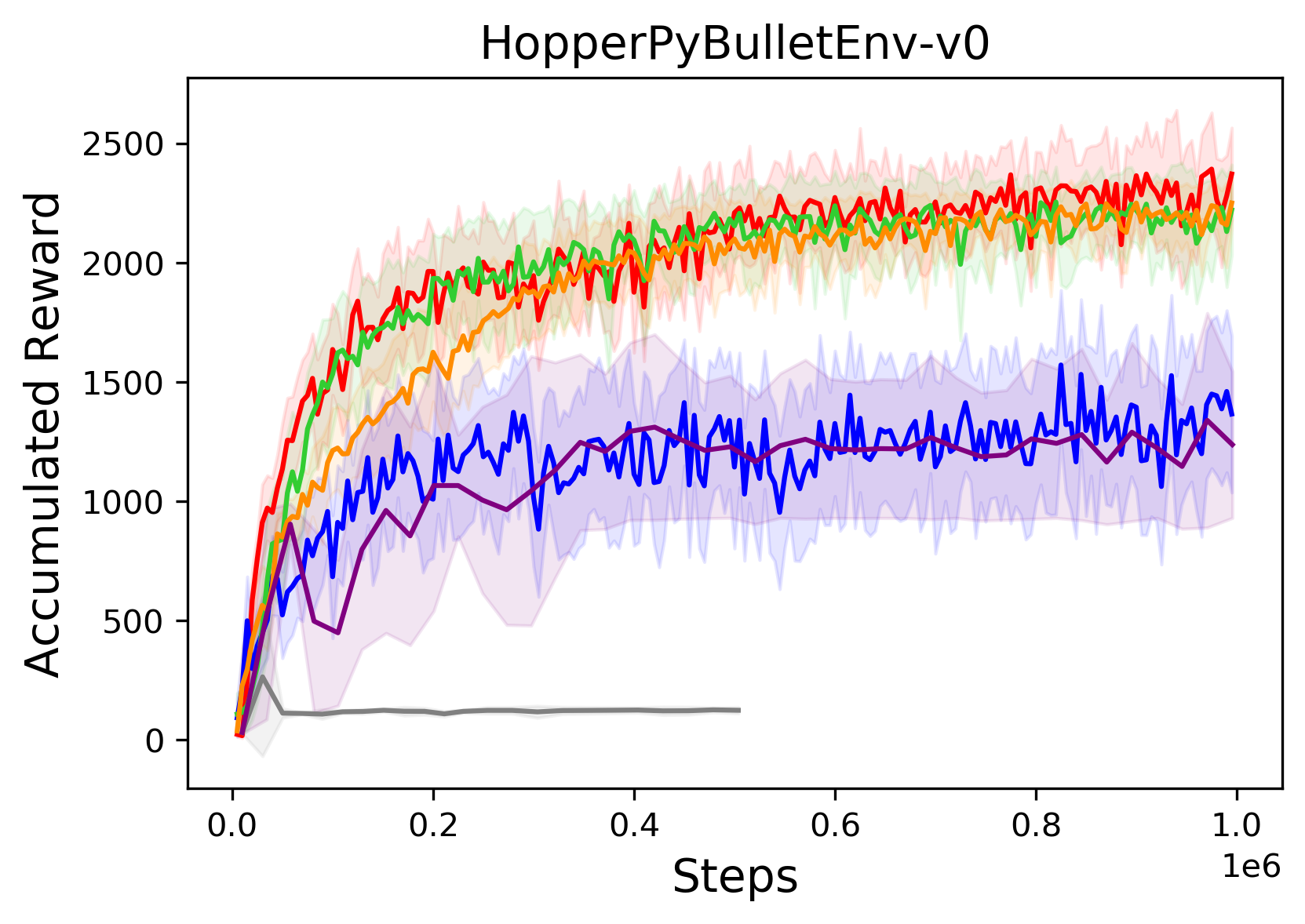}}
    \subfloat[Walker2DPyBulletEnv-v0]{\includegraphics[width=.25\linewidth]{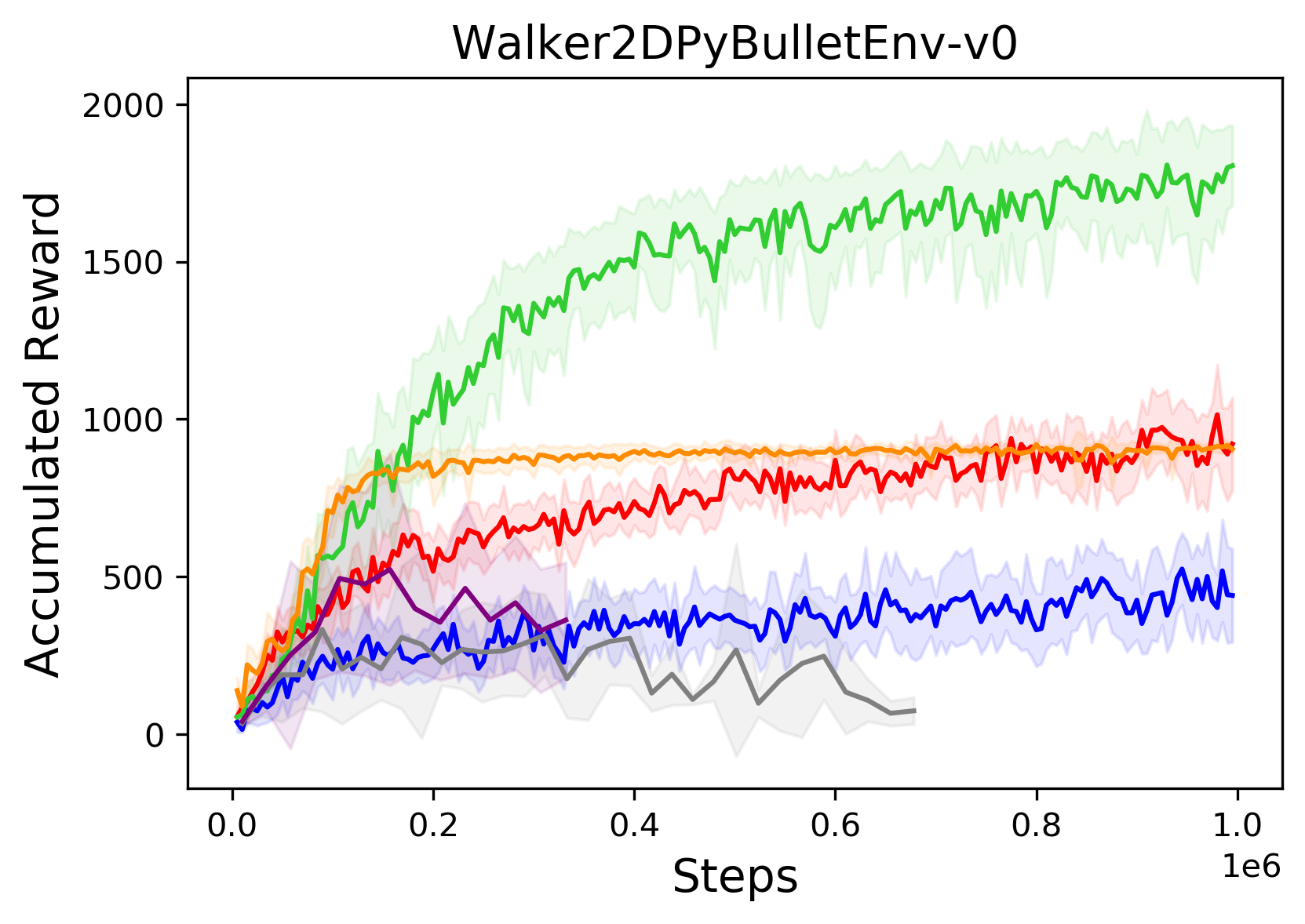}}
    
    \caption{Learning curves for PyBulletGym tasks. The shaded area shows half of standard deviation of the average accumulated return over 10 trails. MEV and STEVE are not run for full 1 million steps as they take more than 58 hours even on a machine with 2 x NVIDIA P100 Pascal and 3 CPUs @ 2.1GHz.}
    \label{fig:learning_curves_for_performance_comparison}
\end{figure*}

\begin{table*}[h]
    \centering
    \caption{Maximum average return over 10 trials of 1 million steps of MMDDPG($8$-avg), DDPG, TD3, SAC, MVE and STEVE. The maximum value for each task is \textbf{bolded}.}
    \label{table:maximum_average_return}
    \small
    \begin{tabular}{rcccccc}\toprule
    \textbf{Tasks} & \textbf{MMDDPG($8$-avg)} & \textbf{DDPG} & \textbf{TD3}  & \textbf{SAC} & \textbf{MVE} & \textbf{STEVE}        \\ \hline
    AntPB         & \textbf{2767.1$\pm$1461.4} & 885.4$\pm$811.7 & 2388.0$\pm$832.6 & 845.5$\pm$103.7 & 639.5$\pm$33.9  & 1969.0$\pm$525.7        \\
    HalfCheetahPB & \textbf{1368.9$\pm$527.3} & 422.2$\pm$137.6 & 1033.9$\pm$429.6 & 608.3$\pm$131.1 & 331.9$\pm$290.6 & 630.5$\pm$132.2          \\
    Walker2dPB    & 1014.0 $\pm$316.3 & 524.0$\pm$227.5 & \textbf{1806.8$\pm$270.0} & 918.9$\pm$33.4 & 332.4$\pm$261.8 & 522.7$\pm$368.3           \\
    HopperPB      & \textbf{2391.9$\pm$473.3} & 1570.7$\pm$626.8 & 2253.9$\pm$295.2 & 2249.9$\pm$207.9 & 263.4$\pm$332.4 & 1338.6$\pm$449.6         \\
    AntMJC        & 3042.8$\pm$1038.5 & 2014.8$\pm$1371.6 & \textbf{3495.2$\pm$725.9} & 1680.5$\pm$414.6 & $-$  &  $-$    \\
    HalfCheetahMJC & \textbf{2242.2$\pm$338.6} & 1311.8$\pm$1367.6 & 2201.1$\pm$692.8 & 1977.7$\pm$180.4 &  $-$ &  $-$    \\ 
    Walker2dMJC   & 1365.6$\pm$409.9 & 844.7$\pm$521.1 & \textbf{1583.3$\pm$670.1} & 779.1$\pm$178.7 &  $-$ &       $-$      \\
    \bottomrule
    \end{tabular}
\end{table*}

This section focuses on comparing MMDDPG($8$-avg) with other baselines namely DDPG, TD3, SAC, MVE and STEVE. Special attention is given to MVE and STEVE, because these two algorithms are very similar to MDDPG and MMDDPG with the difference that they expand multi-steps in a learned environment model. Fig. \ref{fig:learning_curves_for_performance_comparison} shows the learning curves of these algorithms on various tasks, and Table \ref{table:maximum_average_return} summarizes the maximum average return. Obviously, MMDDPG($8$-avg) significantly outperforms DDPG on all examined tasks. Surprisingly, MMDDPG($8$-avg) performs comparably and even better on some tasks than TD3 which is currently one of the state-of-the-art approaches, specifically designed to address function approximation error in DDPG. Considering all examined tasks in this work have a dense reward signal, where multi-step methods' effect on enabling fast propagation of reward will be less important, we speculate that multi-step plays a similar role in alleviating the overestimation problem as does TD3, but using a different mechanism. Compared with MDDPG and MMDDPG, the disadvantage of TD3 is it introduces more computation for training its critics, because it maintains two separate critics and at each training step these two critics are updated to the minimum estimated Q-value of their target critic networks. Detailed comparison in terms of computation cost among DDPG, MDDPG, MMDDPG and TD3 will be discussed in Section \ref{subsec:Computation_Resource_Consumption_Comparison_with_TD3}. 

Counterintuitively, SAC performs worse than TD3 on tasks from PyBulletGym, and for some tasks SAC is even worse than DDPG. This is unexpected, as it is shown in \cite{haarnoja2018soft} that SAC outperforms TD3 on some difficult continuous control tasks from OpenAi gym which use the MuJoCo \cite{todorov2012mujoco} physics engine. One possible explanation is that PyBulletGym uses Bullet physics \cite{coumans2019}, at the same time environments in PyBulletGym are ported from Roboschool environments which are harder than MuJoCo gym, as the robot's body is heavier than that in MuJoCo tasks and termination states are added if robot flips over. This needs further investigation. Especially if SAC is going to be employed in a real robot, the belief that SAC is the best choice might be misleading.

MVE performs the worst on most tasks. STEVE is shown to be sample efficient on AntPyBulletEnv-v0 and is better than MVE, which is consistent with the results in \cite{buckman2018sample}. However, STEVE is worse than MMDDPG($8$-avg) and TD3.

\section{Discussion}

In this section, we discuss the advantages and disadvantages of different ways to do multi-step expansion, and expose the tradeoff between overestimation and underestimation that underlies offline multi-step methods. Then, we compare the computation resource consumption between TD3 and our proposed methods, since they show comparable final performance and learning speed.

\subsubsection{Comparison of Multi-step Expansion Methods}
\label{subsec:Comparison_of_Multi_step_Expansion_Methods}

As shown in Eq. \ref{eq:methods_to_implement_multistep_backup}, there are three ways to calculate $\hat{Q}_t^{(n)}(s_t,a_t)$ depending on how the $n-1$ experiences after $(s_t,a_t,r_t,d_t)$ are acquired: \textbf{(1) offline expansion}, sampled from the replay buffer, e.g. MDDPG and MMDDPG; \textbf{(2) online expansion}, sampled from the environment according to an online policy, e.g. $Q(\sigma)$ \cite{de2018multi}; \textbf{(3) model-based expansion}, sampled from a learned environment model according to an online policy, e.g. MVE and STEVE. 

\begin{equation}
   \hat{Q}_t^{(n)}(s_t,a_t) = r_t + \overbrace{\underbrace{\sum_{k=1}^{n-1}\gamma^{k}r_{t+k}}_{underestimation \ prone}}^{offline,\ online, \ model} + \gamma^{n} \underbrace{\max_{a} Q_{\theta^{Q-}}(s_{t+n}, a)}_{overestimation \ prone}
    \label{eq:methods_to_implement_multistep_backup}
\end{equation}

Theoretically, online expansion is the best as the multi-step experiences are directly sampled from the environment. However, this is unrealistic, because expanding multi-step for each experience $(s_t, a_t, r_t, s_{t+1})$ in a mini-batch is time-consuming, especially when running multiple parallel environments (e.g. in simulation) is impossible. 

A compromise is learning an environment model, then doing multi-step expansion on the learned environment model, as is done in MVE and STEVE. The challenge with this approach is that learning an environment, including transition dynamics and reward function, might be as hard or even harder than learning a policy, even without considering the extra cost for computation resources. It is also not clear to what extent the error introduced by the learned environment model will harm the learning of a policy. As shown in Fig. \ref{fig:learning_curves_for_performance_comparison} and Table \ref{table:maximum_average_return}, MVE and STEVE do not provide significant benefit, compared with MMDDPG($8$-avg) and TD3. 

\begin{figure}[h!]
    \centering
    \includegraphics[width=.85\linewidth]{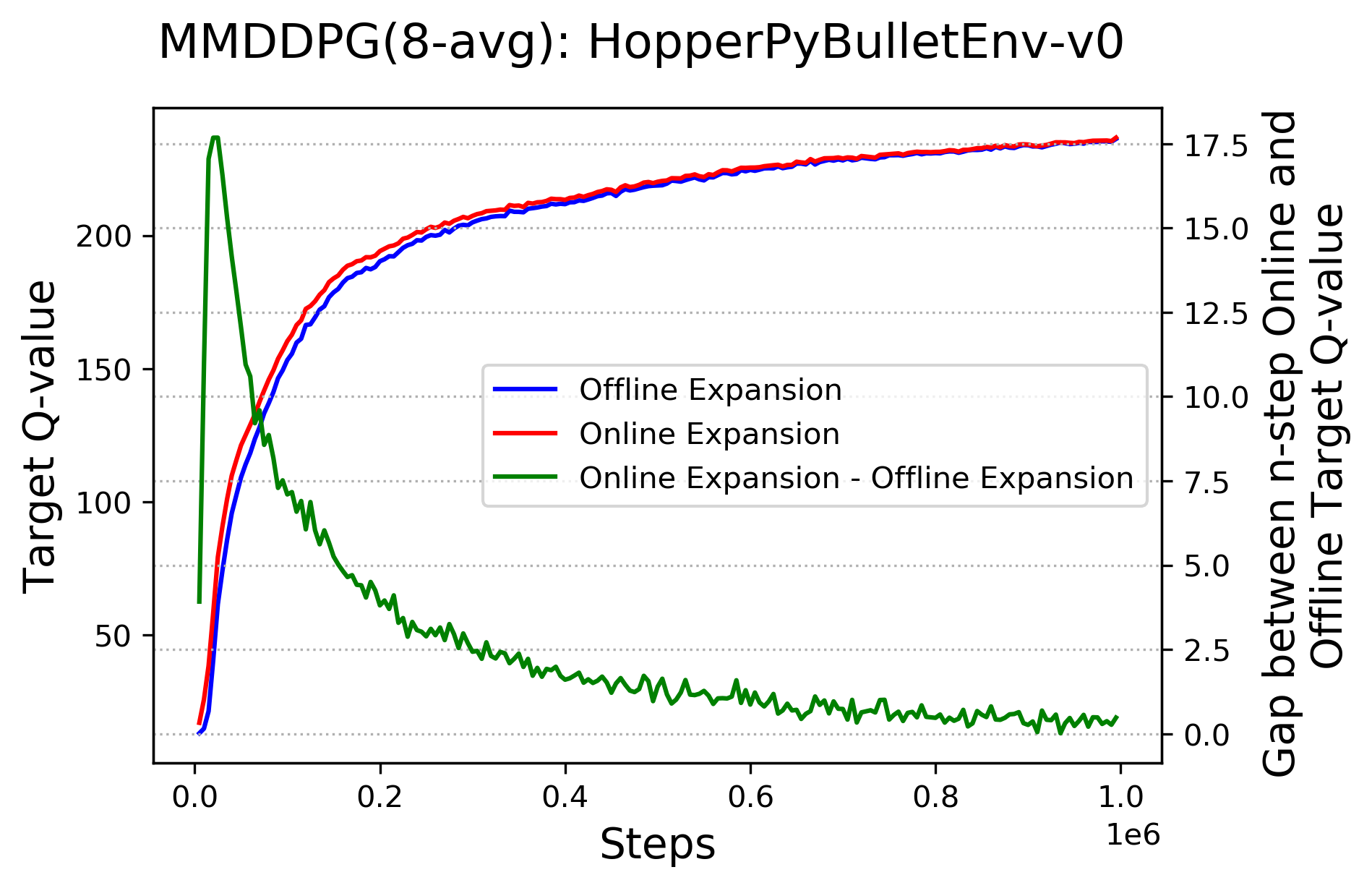}
    \caption{Comparison between Online and Offline Multi-step Expansion, where the blue and the red line correspond to average of offline and online multi-step expansion over a mini-batch sampled from replay buffer, and the green line is the gap between them.}
    \label{fig:Comparison_between_Online_and_Offline_Multi_step_Expansion}
\end{figure}

Offline expansion is a solution somewhat in between. On one hand, it does not need to learn an environment model or to expand according to current policy on the environment, but uses past experiences after $(s_t, a_t, r_t, s_{t+1})$ as an expansion of the current online policy on the environment, with only negligible extra computation required. On the other hand, it is not an exact expansion of online policy, which introduces error in the estimated target Q-value and tends to be an underestimation of Q-value following current online policy. Nevertheless, offline expansion gradually approaches online expansion as the replay buffer fills with experiences following a stable optimal policy. This is seen in Fig. \ref{fig:Comparison_between_Online_and_Offline_Multi_step_Expansion} where the initial gap between online and offline multi-step expansion is big, indicating large underestimation, but gradually decreases with the increase of interactions.

Obviously, multi-step expansion cannot completely overcome the overestimation problem, because the bootstrapped Q-value after $n$-steps is still prone to overestimation as shown in Eq. \ref{eq:methods_to_implement_multistep_backup}. But since $n>1$, the bootstrapped part is weighted less than in the $1$-step method.  Overall, offline multi-step expansion tends to be an underestimation of the online multi-step expansion, while the bootstrapped Q-value after $n$-step tends to be an overestimation of the value in state $s_{t+n}$. Therefore, the step size $n$ balances the tradeoff between overestimation and underestimation.

\subsubsection{Computation Resource Consumption Comparison with TD3}
\label{subsec:Computation_Resource_Consumption_Comparison_with_TD3}

Similar to MVE and STEVE, MDDPG and MMDDPG proposed in this paper employ multi-step expansion to provide a more accurate target Q-value estimation for the critic in DDPG. As discussed in Section \ref{subsec:Performance_Comparison} and \ref{subsec:Comparison_of_Multi_step_Expansion_Methods}, MMDDPG outperforms MVE and STEVE in terms of learning speed, final performance and computation resource consumption. However, unlike multi-step expansion, TD3 takes the minimum of target Q-values estimated from two critics as the final target Q-value to update these two critics, to avoid value approximation error. MDDPG and MMDDPG show comparable final performance and learning speed as TD3 on most tasks.  Here we focus on comparing the computation resource consumption of these three methods. 

\begin{table}[h]
    \centering
    \caption{Comparison of Forward and Backward Propagation on a Mini-bach For Updating the Critic}
    \label{tab:Comparison_of_Forward_and_Backward_Propagation_on_a_Mini_bach_For_Updating_the_Critic}
    \begin{tabular}{c|c|c|c|c}\toprule
       & DDPG & TD3 & MDDPG($n$) & MMDDPG($n$)  \\\hline
    FP &  1 & 2 & 1 & $n$\\
    BP &  1 & 2 & 1 & 1 \\
    Total &  2 & 4 & 2 & $n$+1 \\
    \bottomrule
    \multicolumn{5}{l}{\footnotesize{FP: Forward Propagation, BP: Backward Propagation. }}
    \end{tabular}
\end{table}

Table \ref{tab:Comparison_of_Forward_and_Backward_Propagation_on_a_Mini_bach_For_Updating_the_Critic} summarizes the time of forward and backward propagation needed for training the critic on a mini-batch of transitions. As shown in the table, DDPG needs 1 forward propagation to estimate the target Q-value and 1  backward propagation to update the critic. TD3 performs 2 forward and 2 backwards propagations, one for each critic. MDDPG with a specific step size $n$ does not introduce extra propagations compared with DDPG. However, the number of forward propagations needed for MMDDPG with a specific choice of step size $n$ is $n$, while only 1 backward propagation is needed for the updating critic. Therefore, MDDPG consumes less computation resource than TD3, while MMDDPG consumes more computational resource than TD3 only when $n\geq4$, assuming the forward and backward propagation are equally  demanding in terms of computational resources.

\section{CONCLUSIONS}

In this paper, we empirically revealed multi-step methods' effect on alleviating overestimation in DRL, by proposing MDDPG and MMDDPG which are a combination of DDPG and multi-step methods, and discussed the underlying underestimation and overestimation tradeoff. Results show that employing multi-step methods in DRL helps  to alleviate the overestimation problem by exploiting bootstrapping. This paper also discussed the advantages and disadvantages of three ways to implement multi-step methods from the point of view of extra computation cost and modeling error.

However, a principled way for choosing step size $n$ is still needed. Perhaps dynamically tuning $n$ during the course of learning is more suitable as at different stages of learning the trade-off between overestimation and underestimation needs to be balanced differently. The most important future direction arising from this work is to find a more effective way to overcome overestimation since this is key to improving DRL algorithms' sample efficiency, while still retaining a simple exploration method in order to limit computational needs.

\section*{Acknowledgment}
This research was enabled in part by support provided by SHARCNET (www.sharcnet.ca) and Compute Canada (www.computecanada.ca). The authors would like to thank Daiwei Lin for fruitful discussion. This work is supported by a SSHRC Partnership Grant in collaboration with Philip Beesley Architect, Inc.



\bibliographystyle{IEEEtran}
\bibliography{IEEEabrv,./IEEEexample}

\end{document}